\documentclass{bmvc2k}
\usepackage{comment}
\usepackage{subcaption} % Removed subfigure import from line 148 of .cls
\usepackage{transparent}
\usepackage{amssymb}

\title{Feature Splatting for Better Novel View Synthesis with Low Overlap}

\addauthor{Tomas Berriel Martins}{tberriel@unizar.es}{2}
\addauthor{Javier Civera}{jcivera@unizar.es}{2}

\addinstitution{
 I3A\\ % i3A o Ropert?
 University of Zaragoza\\
 Zaragoza, Spain
}

\runninghead{Berriel Martins, Civera}{Feature Splatting}

\begin{document}

\maketitle

\begin{abstract}
3D Gaussian Splatting has emerged as a very promising scene representation, achieving state-of-the-art quality in novel view synthesis significantly faster than competing alternatives. 
However, its use of spherical harmonics to represent scene colors limits the expressivity of 3D Gaussians and, as a consequence, the capability of the representation to generalize as we move away from the training views. In this paper, we propose to encode the color information of 3D Gaussians into per-Gaussian feature vectors, which we denote as Feature Splatting (FeatSplat). To synthesize a novel view, Gaussians are first "splatted" into the image plane, then the corresponding feature vectors are alpha-blended, and finally the blended vector is decoded by a small MLP to render the RGB pixel values. To further inform the model, we concatenate a camera embedding to the blended feature vector, to condition the decoding also on the viewpoint information.
Our experiments show that these novel model for encoding the radiance considerably improves novel view synthesis for low overlap views that are distant from the training views. Finally, we also show the capacity and convenience of our feature vector representation, demonstrating its capability not only to generate RGB values for novel views, but also their per-pixel semantic labels. Code is available on the \href{https://tberriel.github.io/projects/FeatSplat/}{project page}.


\end{abstract}
%-------------------------------------------------------------------------
\section{Introduction}
\label{sec:intro}
Finding appropriate 3D scene representations is a key challenge in order to enable robotics, VR and AR applications. In general, representations should not only encode a scene's geometry. Radiance properties are very relevant for novel view synthesis, and high-level aspects such as semantics or affordances are essential to define robotics tasks. In the last years, Neural Radiance Fields (NeRFs)~\cite{mildenhall2021nerf} have been the dominant approach in the research literature to learn implicit scene representations from images. However, NeRFs are computationally demanding at training and rendering and scale badly with the size of the scene, which has limited their use in practical applications. 

3D Gaussian Splatting (3DGS)~\cite{kerbl20233d} appeared very recently as a promising alternative to NeRFs. 3DGS is able to achieve similar quality for novel view synthesis significantly faster, although with a much higher memory footprint. Representing a scene with 3D Gaussians, however, does not come without limitations. As the most relevant to our work, a explicit scene representation composed of 3D Gaussians and spherical harmonics lacks the capacity to model complex image textures. This limitation is typically alleviated by an increase of the number of Gaussians and a reduction of their size, so that each tiny Gaussian models a smaller and simpler part of the texture. Such strategy, in addition to being intensive in memory, leads to overfitting to the training views and to a poor generalization.
% Explain it's main limitations

\begin{figure}
\begin{center}
\def\svgwidth{0.98\linewidth}
    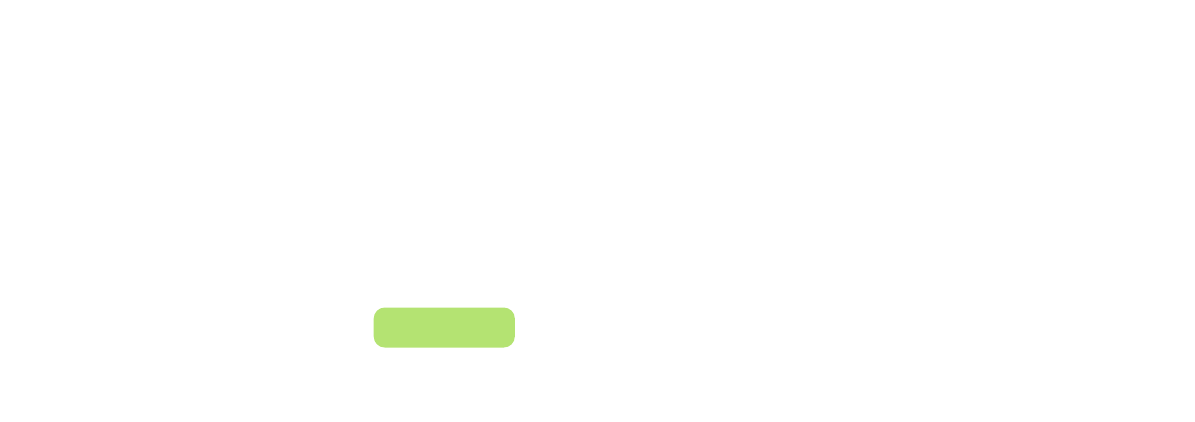
\end{center}
    \caption{Overview of our approach. \textit{Left:} We augment the 3D Gaussian Splatting~\cite{kerbl20233d} representation with learned feature vectors \(\mathbf{f}\) to encode colors, removing the spherical harmonics. \textit{Center:} To render a point of view, 3D Gaussians are projected to the image plane, where the differentiable rasterizer alpha-blends corresponding feature vectors. Resulting vectors are concatenated with a camera embedding, and a tiny MLP renders the final RGB values, with potentially also higher level information such as semantic labels. \textit{Right}: Illustrative result of novel view synthesis result with low overlap, with RGB values and semantic labels.}
    \label{fig:our_method}
\end{figure}

In this paper we propose a novel approach, that we denote as Feature Splatting (FeatSplat), that replaces spherical harmonics by a feature vector representation for each 3D Gaussian. Using such representations, we show in ScanNet++~\cite{yeshwanthliu2023scannetpp} a significant improvement in novel view synthesis with low overlap. This comes as a result of the higher capacity of our feature vector representation compared to spherical harmonics. Figure \ref{fig:our_method} illustrates our method. We splat the 3D Gaussians and alpha-blend the feature vectors. We then concatenate to the splatted and blended feature vector dimensions to camera intrinsics and extrinsics channels, and finally decode them into colors with a MLP. As an additional use case illustrating the flexibility of our representation, we also show its capability to render accurate semantic labels. 
\section{Related Work}
% Implicit representations (NeRFs)
In the last years, fully implicit scenes representations have been typically modelled as a function that maps scene coordinates to the desired scene attributes (\emph{e.g.}, color), and a neural network is trained on the scene data so that it learns such function. One of the most successful examples are NeRFs~\cite{mildenhall2021nerf}. 
A NeRF consists on a Multi Layer Perceptron (MLP) that maps a 3D point and a viewing direction to a RGB color and an occupancy or density value, coupled with a differentiable rendering equation to aggregate values of 3D points along a ray casted through each pixel to compute the final RGB color. Given a set of posed images, the MLP is optimized to perform novel view synthesis afterwards.
To compensate for the initial high training and rendering times, several approaches have been developed. Some proposed to use depth information to reduce sampling in 3D~\cite{deng2022depth,wei2021nerfingmvs,neff2021donerf, roessle2022dense}; others combined the MLP with explicit 3D representations like Octrees~\cite{yu2021plenoctrees,fridovich2022plenoxels}, Voxels~\cite{liu2020neural}, 3D point clouds~\cite{xu2022point} and 3D hierarchical mesh grids~\cite{muller2022instant} and encoded into it learned features that where then quickly decoded using a smaller MLP.

Due to its ability to encode the scenes' radiance and geometry, the original NeRF formulation has been extended with other capabilities and representations like depth estimation~\cite{li2021mine, guizilini2022depth,guizilini2023delira,guizilini2023towards}; Signed Distance Functions (SDFs)~\cite{wang2021neus,neus2}; semantics, for both panoptic~\cite{kundu2022panoptic} and open vocabulary~\cite{kerr2023lerf} segmentation; and have been also integrated in a wide variety of use cases like scene editing~\cite{wang2023dmnerf, zhang2021editable}, robotics~\cite{shen2023distilled}, and SLAM~\cite{sucar2021imap, zhu2022nice, sandstrom2023point}.

% Explicit representations (3D Gaussian Splatting and its variants)
Recently, 3DGS~\cite{kerbl20233d} has emerged as a substantially faster yet still high-quality alternative. It relies on the explicit modelling of the 3D geometry by a set of 3D Gaussians, each of them with a set of spherical harmonics (SHs)~\cite{yu2021plenoctrees} to represent their color. Rather than using ray tracing like NeRFs, Gaussians are rasterized by being projected into the image plane, sorted by the Radix algorithm~\cite{merrill2010revisiting}, and combined using alpha-blending to give the RGB color for every pixel. The rendered image is then used to compute a photometric loss, and the Gaussians' parameters are optimized using backpropagation. By avoiding the  MLP training and relying in an efficient sorting algorithm and CUDA implementation of the rasterizer, 3DGS is the state of the art for the combined rendering performance, training time, and inference rendering speed. 
Nevertheless, 3DGS is also characterized by 1) a high memory usage; 2) poor surfaces' reconstructions; and 3) the limited capacity of SHs, which increases the number of Gaussians needed for high quality results, worsening the previous two problems.
Several approaches augment Gaussians adding latent vectors which are latter decoded. 
LangSplat~\cite{qin2023langsplat} combines CLIP~\cite{radford2021learning} features with SAM~\cite{kirillov2023segany} segmentation masks to generate a language field enabling open-vocabulary 2D semantic segmentation. Similarly, both Feature 3DGS~\cite{zhou2024feature} and Gaussian Grouping~\cite{gaussian_grouping}, use SAM~\cite{kirillov2023segany} as a teacher to embed in each Gaussian a semantic feature. FMGS~\cite{zuo2024fmgs} uses a hash grid based on iNGP~\cite{muller2022instant} to encode CLIP and DINOv2~\cite{oquab2023dinov2} features, to also generate an intrinsic representation capable of open-vocabulary 2D segmentation. 
Nevertheless, none of them neither address the SHs capacity nor show improvement in novel view synthesis, as we do.
Based also on iNGP, Compact-3DGS~\cite{lee2023compact} adds a hash grid that encodes a latent vector for each Gaussian and uses a small MLP conditioned by the viewing direction to recover the corresponding color for each point-of-view. 
Similarly, our FeatSplat relies on using a latent vector to encode each Gaussian's color. But, differently from it, we directly attach a feature vector to each Gaussian, and, during the rasterization, rather than first decoding the colors and then alpha-blending, we first alpha-blend the feature vectors of corresponding Gaussians, and then decode the final vector to RGB, which allows us to condition on the intrinsic and extrinsic camera parameters. In our experiments, we show that FeatSplat, due to these differences, outperforms Compact-3DGS at novel view synthesis.
\section{Preliminaries: 3D Gaussian Splatting}
3DGS~\cite{kerbl20233d} is an explicit scene representation that relies on a set of 3D Gaussians to encode a scene, which is optimized by differentiable rasterization from known viewpoints. 

\paragraph{Representation.} The geometry of a 3D Gaussian~\(G\) is parameterized by its 3D center position \(\mathbf{x}_{G} \in \mathbb{R}^3\) and a 3D covariance $\Sigma_G= R_G S_G S_G^\top R_G^\top \succeq 0$ that can be decomposed into a rotation matrix \(R_G \in SO(3)\) and a scaling matrix  \(S_G \in \text{diag}(\mathbb{R}^3_{\geq 0})\). Each Gaussian has a density value \({\sigma}_G \in \mathbb{R}\); and a set of spherical harmonics (SHs) coeficients \(\mathbf{k}_G \in \mathbb{R}^h\) that parameterize the function \(\mathcal{S}_{\mathbf{k}_G}\left(\mathbf{d}_G\right)\) to encode RGB values \(\mathbf{c}_G \in \mathbb{R}^3\) depending on the viewing direction \(\mathbf{d}_G \in S^2\) relative to each Gaussian.

\paragraph{Differentiable rendering.} To render a RGB image \(\hat{{I}} \in \mathbb{R}^{w \times h \times 3}\) of the scene, first SHs are converted to RGB, \(\mathbf{c}_G = \mathcal{S}_{\mathbf{k}_G}\left(\mathbf{d}_G\right)\). Then,  Gaussians are projected, or splatted, into 2D Gaussians in the image plane~\cite{zwicker2001ewa} and sorted using the Radix algorithm. Finally, the color \(\textbf{c}_p\) for each pixel \(\mathbf{p} \in \Omega\), \(\Omega\) standing for the image domain, comes from the alpha blending of the \(\mathcal{N}\subset G \) 2D Gaussians falling into it
\begin{equation}
\label{eq:color}
    \textbf{c}_p = \sum_{i\in\mathcal{N}}{T_i \alpha_i \mathbf{c}_i} \text{, with } T_i =\prod_{j=i}^{i-1}\left(1-\alpha_i\right),
\end{equation}
where \(\alpha_i \in \mathbb{R}\) is obtained by multiplying the learned occupancy value \(\sigma_i\) by the evaluation of the projected 2D Gaussian~\cite{yifan2019differentiable, kopanas2021point,kopanas2022neural} for each pixel \(\mathbf{p}\).

\paragraph{Optimization.} The representation is fitted as follows. The 3D Gaussians are initialized either randomly or from a sparse Structure from Motion point cloud. At each optimization step, the Gaussians are used to render a viewpoint from the training set and compute the loss
\begin{equation}
    \mathcal{L} = \left(1-\lambda_{SSIM}\right)\mathcal{L}_1\left(\hat{{I}},{I}\right) + \lambda_{SSIM} \mathcal{L}_{D-SSIM}\left(\hat{{I}},{I}\right),
\end{equation}
between the rendered image \(\hat{{I}}\) and the ground truth \(I\). \(\mathcal{L}_1(\cdot,\cdot)\) stands for the L${_1}$-norm of the difference of its two arguments, \(\mathcal{L}_{D-SSIM}(\cdot,\cdot)\) is the Structure Similarity Index Measure (SSIM)~\cite{ssim} between its two arguments, and \(\lambda_{SSIM}\) weights both terms.
Adam~\cite{Kingma2014AdamAM} is used to update the Gaussians' parameters by backpropagating the loss, and Gaussians are pruned or cloned adaptively according to certain policies.

Despite SHs encoding direction-dependent RGB values, their variability is limited, and are only able to encode one color per viewing direction. To represent complex textures in natural scenes, 3DGS over-densifies scenes by optimizing Gaussians at different depths, visible only from certain viewpoints. In a manner, this strategy abuses the geometric primitive to overfit the photometry, achieving high-quality rendering for views close to the training distribution, but causing artifacts observable from viewpoints far from the training views.
\section{Feature Splatting (FeatSplat)}
Our FeatSplat relies on learned feature vectors instead of SHs to better encode information and increase the capacity and flexibility of 3D Gaussians.
Each 3D Gaussian is initialized with a feature vector \(\mathbf{f}_G\) sampled from a normal distribution \(\mathcal{N}\left(0,1\right)\). When rendering an image \(\hat{I}\), the \(\mathcal{N}\) \(\mathbf{f}_G\) vectors splatted in a pixel \(\mathbf{p}\) are alpha-blended
\begin{equation}
    \textbf{f}_p = \sum_{i\in\mathcal{N}}{T_i \alpha_i \mathbf{f}_i},
\end{equation}
similarly to what was done with colors in Eq.~\eqref{eq:color}. After that, a small MLP transforms \(\textbf{f}_p\), into the corresponding RGB color \(\textbf{c}_p\) of the target image. The per-pixel feature vectors \(\textbf{f}_p\) are concatenated with camera parameter embeddings, in order to condition the MLP not only on the information of the 3D Gaussians, but also on the viewpoint. The camera extrinsics are represented by the 3D camera position of the target image in global coordinates, and the intrinsics by an embedding of the relative position of the rendered pixel in the target image plane. The final rendered color is then
\begin{equation}
    \textbf{c}_p = \text{MLP}\left(\textbf{f}_p\oplus \mathbf{x}_{cam} \oplus \textbf{e}_p \right),
\end{equation}
where \(\oplus\) is a concatenation operation, \(\mathbf{x}_{cam} \in \mathbb{R}^3\) is the 3D camera position, and \(\mathbf{e}_p \in [-1,1] \times [-1,1]\) the two-dimensional pixel embedding.
%Although the value of the 2D pixel encoding will depend from the camera resolution and pixel position in the image, which are intrinsic parameters, given that the MLP evaluates each pixel independently
We also considered adding camera orientation information, but our ablations (App.~\ref{sec:ablation}) did not show a benefit.

\paragraph{Semantic segmentation.} 
Learning per-Gaussian feature vectors \(\textbf{f}_p\) increases the capability of a 3DGS representation. In this paper we illustrate how FeatSplat can be adapted for 2D semantic segmentation. All is needed is changing the output dimension of the MLP from \(3\) channels to \(3 + \#\text{classes}\) and training with an additional semantic cross-entropy loss \(\mathcal{L}_{ce}\).
\begin{equation}
    \mathcal{L} = \left(1-\lambda\right)\mathcal{L}_1\left(\hat{\text{I}},\text{I}\right) + \lambda \mathcal{L}_{D-SSIM}\left(\hat{\text{I}},\text{I}\right) + \lambda_{sem} \mathcal{L}_{ce}\left(\hat{S},S\right),
\end{equation}
where \(\lambda_{sem}\) is a hyperparameter, and \(\hat{S}\) and \(S\) are the predicted and ground truth segmentation labels. 
\begin{figure}[h]
\begin{center}
\includegraphics[width=\linewidth]{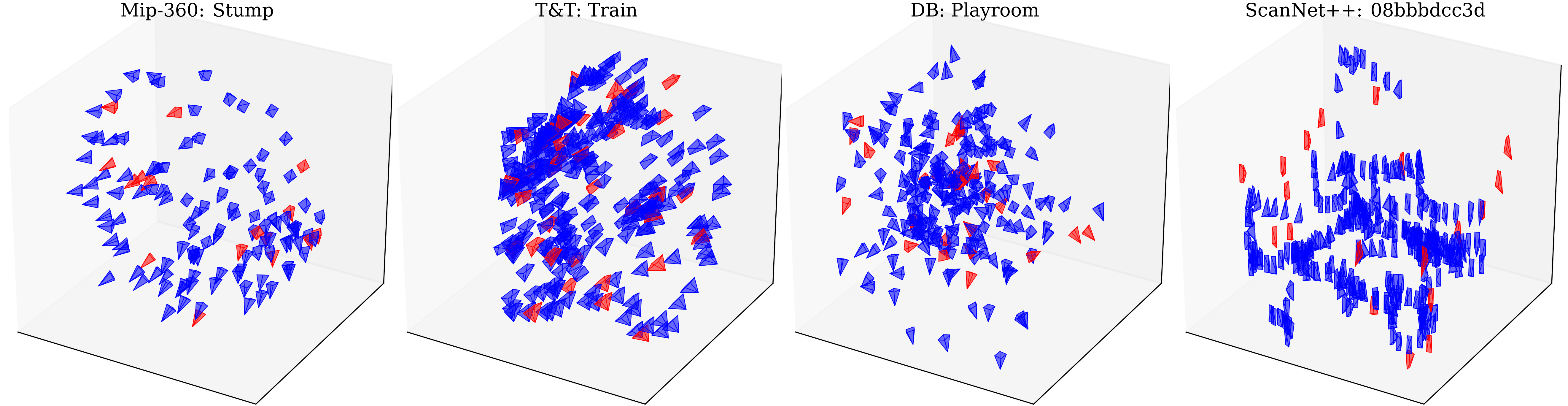}
\end{center}
    \caption{Comparison of 3D position of test (red) and training (blue) points of view on one scene for each of the four datasets.}
    \label{fig:cam_dist}
\end{figure}
\section{Experiments}
\paragraph{Implementation details.} We implemented FeatSplat with both $16-$ and $32-$dimensional feature vectors for novel view synthesis, and only $32-$dimensional ones for novel view synthesis plus semantic segmentation. For each scene, we train a MLP with one hidden layer of 64 neurons, common for all Gaussians. For semantic segmentation, the model is trained on 64 classes: the 63 most frequent classes in ScanNet++, plus an unknown category label. Further implementation and optimization details are included in App.~\ref{app:implementation}.

\paragraph{Baselines.} We compare FeatSplat against 3DGS~\cite{kerbl20233d}, the current state of the art for real-time novel view synthesis, Compact-3DGS~\cite{lee2023compact} which also augmented 3DGS with learned feature vectors instead of SHs, and iNGP~\cite{muller2022instant}, which was the previous state of the art. Finally, we also include Zip--NeRF and Mip-NeRF 360 metrics on the datasets they have been reported, as a reference. Nevertheless, FeatSplat should not be compared against them due to their offline nature (they both render at less than 1 frame per second).
Finally, on the semantic segmentation experiment, we do not compare against other models that learn specific semantic features for open-vocabulary semantic segmentation. Given that our model is trained on a closed-vocabulary, we would have an unfair advantage.
\begin{figure}
\begin{center}
    \includegraphics[width=\linewidth]{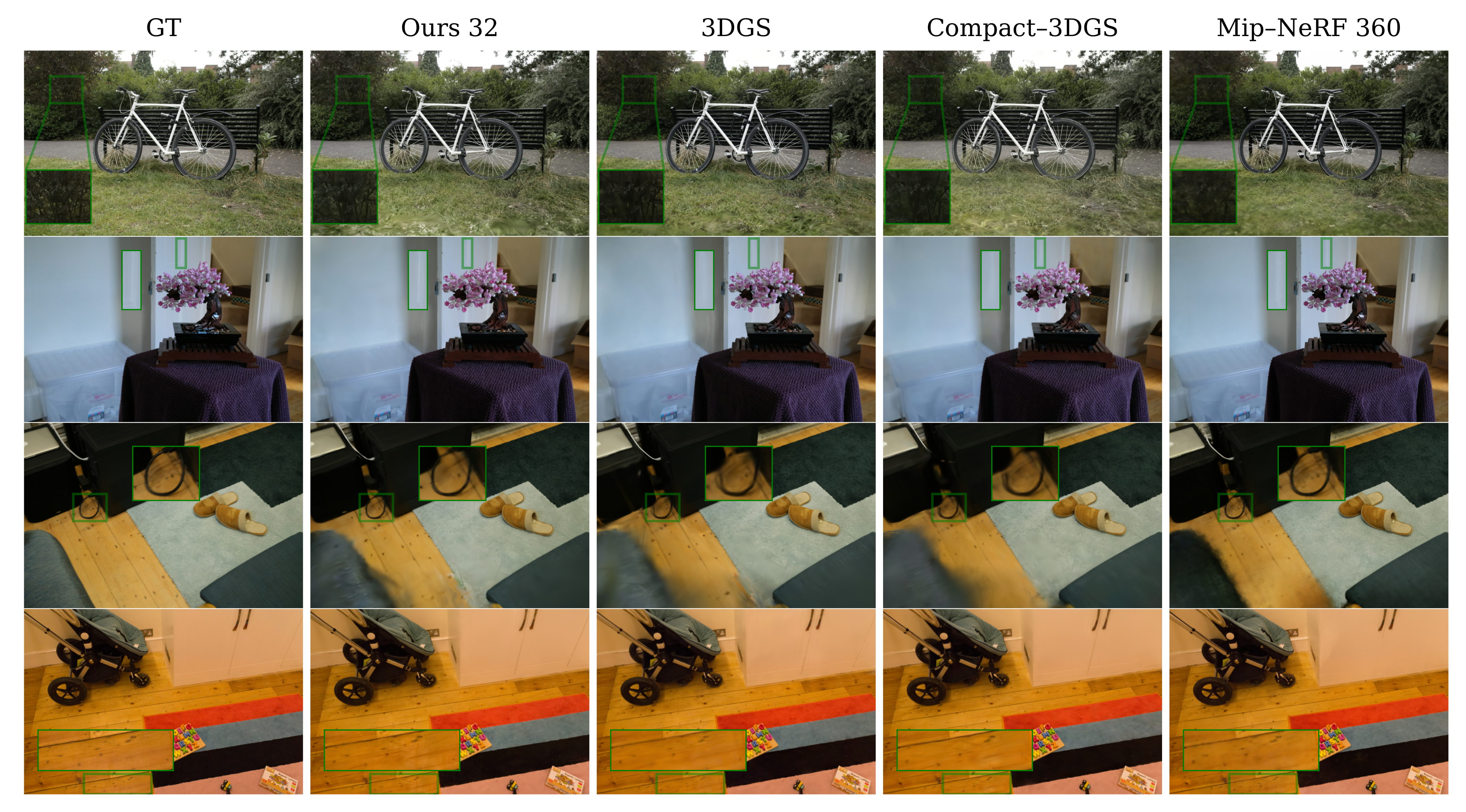}
\end{center}
    \caption{Comparison of our method trained with features with 16 (FeatSplat--16) and 32 dimensions (FeatSplat--32), against prior work. From top to down the scenes are \textit{Bicycle}, \textit{Bonsai}, and \textit{Room} from Mip-360~\cite{barron2022mipnerf360}, and \textit{Playroom} from DB~\cite{hedman2018deep}.}
    \label{fig:gs_dataset}
\end{figure}
\paragraph{Datasets and metrics.} Rendering performance is evaluated using PSNR, LPIPS~\cite{Zhang_2018_CVPR}, and SSIM~\cite{ssim}, computational aspects by the representation size in megabytes (MB), rendering speed in Frames per Second~(FPS) on a GTX 3090, and semantic segmentation by the mIOU weighted over classes' support among all scenes.
We evaluate on three common datasets in the literature~\cite{kerbl20233d,lee2023compact, barron2022mipnerf360}: Mip\-360~\cite{barron2022mipnerf360}, Tanks and Temples~\cite{knapitsch2017tanks} (T\&Ts); and Deep Blending~\cite{hedman2018deep} (DB), for a total of 12 scenes. In these datasets the test set is selected by sampling one out of every eight frames of the image sequences, and the other 7 frames are used to train the model.
We also evaluate FeatSplat in a subset of 21 scenes from the novel ScanNet++~\cite{yeshwanthliu2023scannetpp}. ScanNet++ is a large--scale dataset with high-resolution RGB and semantic annotations of indoor scenes, and a test set generated from an independent camera trajectory specifically designed to evaluate novel view synthesis with low overlap and separated views.

The independent test set from ScanNet++ represents a considerable increase in difficulty with respect to commonly used datasets. Sampling one out of 8 frames from a video stream (as done in Mip-360, T\&Ts and DB) means that at least the previous and following frames would have a significant overlap with the test frames. In contrast, an independent sequence from a different trajectory results in test images looking at unexplored parts of the scene or at the same areas but from rather different viewpoints. In Figure~\ref{fig:cam_dist} we can visualize, for each of the four datasets, the scene with the highest average angular distance from each test viewpoint (red) to its 10 closest training viewpoints (blue). The contrast between datasets is clear. While on Mip-360, T\&Ts, and DB, cameras tend to be nearby each other and looking in the same direction, on ScanNet++ there are several test views that are isolated and looking at mostly unexplored areas.
Further details on the datasets can be found in App.~\ref{app:datasets}.
\begin{figure}
\begin{center}
\includegraphics[width=\linewidth]{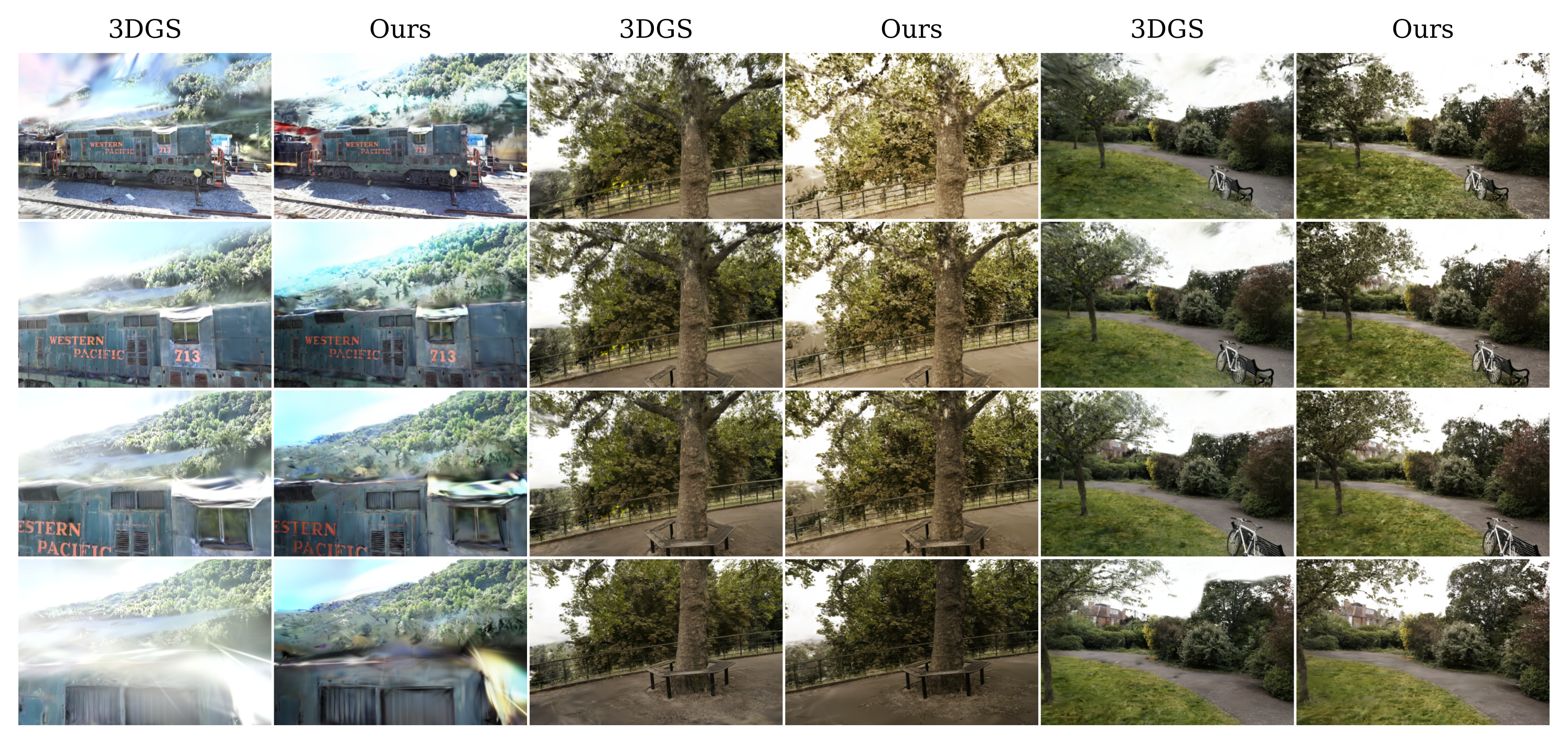}
\end{center}
    \caption{Novel view synthesis with low overlap of FeatSplat--32 against 3DGS. From left to right the scenes are \textit{Train} from T\&Ts~\cite{knapitsch2017tanks}, \textit{Treehill}, and \textit{Bicycle} from Mip-360~\cite{barron2022mipnerf360}. }
    \label{fig:gs_dataset_2}
\end{figure}

\begin{table}[h]
\begin{center}
\resizebox{0.98\linewidth}{!}{
\begin{tabular}{lccccccccccccccc}
\multicolumn{1}{c}{\textbf{}} & \multicolumn{5}{c}{Mip-360}  & \multicolumn{5}{c}{Tanks and Temples} & \multicolumn{5}{c}{Deep Blending} \\
\multicolumn{1}{l|}{}  & SSIM      & PSNR      & LPIPS     & FPS & \multicolumn{1}{l|}{Mem}  &  SSIM      & PSNR      & LPIPS     & FPS & \multicolumn{1}{l|}{Mem}  & SSIM      & PSNR      & LPIPS     & FPS & Mem  \\ \hline
\multicolumn{1}{l|}{INGP--Big}  & 0.699    & 25.59   & 0.331   & 9.4  &  \multicolumn{1}{r|}{48MB}& 0.745 & 21.92  & 0.305  & 14.4  & \multicolumn{1}{r|}{48MB}   & 0.817 & 24.96 & 0.390 & 2.8  & 48MB  \\
\multicolumn{1}{l|}{3D Gaussian Splatting} & \cellcolor[HTML]{FF9D9A}0.813 & \cellcolor[HTML]{FFFC9E}27.39 & \cellcolor[HTML]{FF9D9A}0.218 & 115   & \multicolumn{1}{r|}{880MB}    &  \cellcolor[HTML]{FFCE93}0.844 & \cellcolor[HTML]{FFFC9E}23.65 & \cellcolor[HTML]{FF9D9A}0.179 & 156                   & \multicolumn{1}{r|}{431MB}  & 0.899                         & 29.46                         & \cellcolor[HTML]{FF9D9A}0.247 & 126                   & 662MB \\
\multicolumn{1}{l|}{Compact-3DGS} & 0.798 & 26.97 & 0.244  & 140 &  \multicolumn{1}{r|}{49MB} & 0.831                         & 23.32                         & 0.201                         & 185                     & \multicolumn{1}{r|}{39MB} & \cellcolor[HTML]{FFFC9E}0.901 & \cellcolor[HTML]{FFCE93}29.79 & \cellcolor[HTML]{FFFC9E}0.258 & 181                     & 43MB \\
%\multicolumn{1}{l|}{TRIPS} & 0.748 & 25.216 & 0.192 &  &  &                        &                          &                         &                      & &  &  &  &                      &  \\
\multicolumn{1}{l|}{\textbf{FeatSplat--16 (Ours)}}                                                          & \cellcolor[HTML]{FFCE93}0.808 & \cellcolor[HTML]{FFCE93}27.48 & \cellcolor[HTML]{FFFC9E}0.230 & 96                    & \multicolumn{1}{r|}{275MB} & \cellcolor[HTML]{FF9D9A}0.848 & \cellcolor[HTML]{FFCE93}24.48 & \cellcolor[HTML]{FFFC9E} 0.185 & 171                   & \multicolumn{1}{r|}{149MB}  & \cellcolor[HTML]{FF9D9A}0.904 & \cellcolor[HTML]{FFFC9E}29.78 & \cellcolor[HTML]{FFCE93}0.249 & 111                    & 212MB \\
\multicolumn{1}{l|}{\textbf{FeatSplat--32 (Ours)}}                                                          & \cellcolor[HTML]{FFCE93}0.808 & \cellcolor[HTML]{FF9D9A}27.53 & \cellcolor[HTML]{FFCE93}0.227 & 82                    & \multicolumn{1}{r|}{420MB} & \cellcolor[HTML]{FF9D9A}0.848 & \cellcolor[HTML]{FF9D9A}24.53 & \cellcolor[HTML]{FFCE93}0.183 & 147                   & \multicolumn{1}{r|}{227MB}  & \cellcolor[HTML]{FF9D9A}0.904 & \cellcolor[HTML]{FF9D9A}29.85 & \cellcolor[HTML]{FF9D9A}0.247 & 98 & 327MB \\  \hline
\multicolumn{1}{l|}{Mip--NeRF 360}                                                                  & 0.792                         & 27.29 & 0.237                         & 0.06                    & \multicolumn{1}{r|}{9MB} & 0.759                         & 22.22                         & 0.257                         & 0.14                    & \multicolumn{1}{r|}{9MB}  & 0.901                         & 29.4                          & 0.245 & 0.09                    & 9MB    \\
\multicolumn{1}{l|}{Zip\-NeRF}                                                                  & 0.836                         & 28.54                         & 0.177                         & 0.25                    & \multicolumn{1}{c|}{--} & \multicolumn{5}{c|}{--}   & \multicolumn{5}{c}{--}    

\end{tabular}}
\end{center}
\caption{Evaluation on Mip-360~\cite{barron2022mipnerf360}, T\&Ts~\cite{knapitsch2017tanks} and DB~\cite{hedman2018deep} for novel view synthesis.}
\label{tab:mip360}
\end{table}
\subsection{Novel View Synthesis Results}

\begin{figure}
\begin{center}
\includegraphics[width=\linewidth]{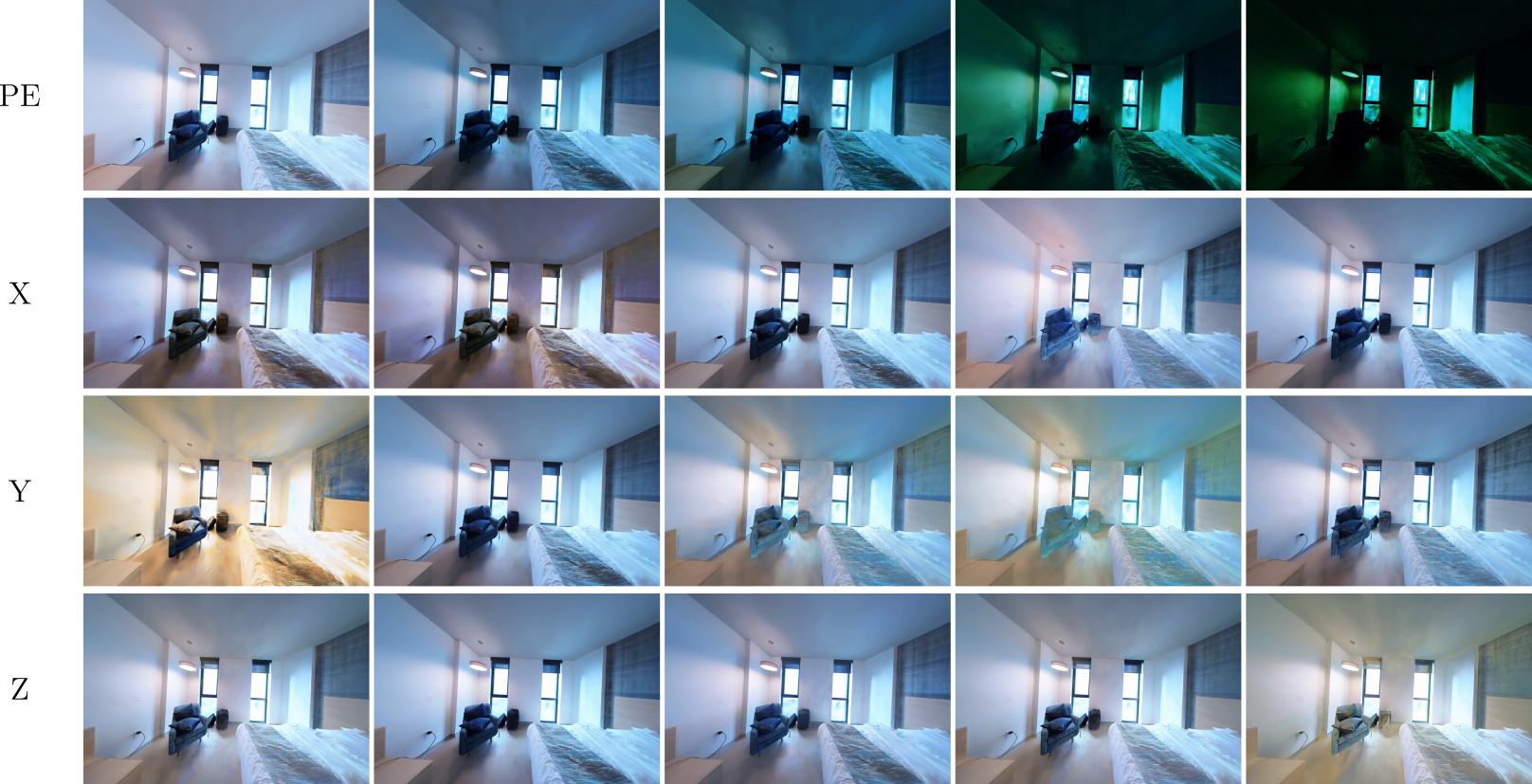}
\end{center}
    \caption{Global illumination change modifying MLP input values of the pixel embedding (\textit{PE}), and the 3D coordinates (\textit{X, Y, Z}), on scene \textit{03f7a0e617} from \textit{ScanNet++}.}
    \label{fig:light_change}
\end{figure}
\textbf{Mip-360, T\&Ts and DB.}
The quantitative results on Mip-360 dataset, T\&Ts and DB in Table~\ref{tab:mip360} show how both FeatSplat models achieve similar rendering performance.
Although our 32-dimensions variant has a slight advantage on rendering quality (equal or better PSNR, LPIPS and SSIM), the 16-dimensional version has higher rendering speed and a smaller representation size. We conclude that 16 dimensions is enough to encode a scene's RGB information.
Comparing FeatSplat against 3DGS and Compact-3DGS, our method consistently achieves the best PSNR on all three datasets, with a better SSIM on two of them. Nevertheless, 3DGS has a better LPIPS overall.
The qualitative results in Figure~\ref{fig:gs_dataset} show how all methods render high quality images, although with slight differences. 
3DGS generates more accurate high-frequency details due to its tendency to over-densify scenes, e.g. the grass on the \textit{Bicycle} scene and the trees on the \textit{Train} scene. 
Instead, FeatSplat--32 is also able to capture details missed by 3DGS (e.g. the bush in \textit{Bicycle}, the door in \textit{Bonsai}, and the floor in \textit{Playroom}) while reducing the amount of artifacts thanks to its more compact and less overfitted representation. This compactness can also be observed in the memory used, that is half for FeatSplat--32 and a fourth for FeatSplat--16. It can be noticed that the rendering speed is reduced for our FeatSplat in comparison to 3DGS and Compact-3DGS. However, we are still well within real-time limits, so we believe this is not a significant limitation.
\begin{table}[h!]
\small
\begin{center}
\begin{tabular}{lcccccc}\multicolumn{1}{c}{\textbf{}}                   & \multicolumn{5}{c}{ScanNet++}                                                                       &  \\
\multicolumn{1}{l|}{}                           & SSIM           & PSNR           & LPIPS          & FPS          & \multicolumn{1}{c|}{Mem}          &  \\ \hline
\multicolumn{1}{l|}{3D Gaussian Splatting}      & \cellcolor[HTML]{FFCE93}0.867          & \cellcolor[HTML]{FFCE93} 23.83          & \cellcolor[HTML]{FFCE93} 0.250          & \textbf{177} & \multicolumn{1}{c|}{217}          &  \\
\multicolumn{1}{l|}{Compact 3DGS}      & \cellcolor[HTML]{FFFC9E} 0.854  & \cellcolor[HTML]{FFFC9E} 22.76    &\cellcolor[HTML]{FFFC9E}  0.272   & 170 & \multicolumn{1}{c|}{\textbf{35}}          &  \\
\multicolumn{1}{l|}{\textbf{FeatSplat--16 (Ours)}}          & \cellcolor[HTML]{FF9D9A}0.875 & \cellcolor[HTML]{FF9D9A}24.62 & \cellcolor[HTML]{FF9D9A}0.245 & 82           & \multicolumn{1}{c|}{74} &  \\
\multicolumn{1}{l|}{\textbf{FeatSplat--32 (Ours)}}          & \cellcolor[HTML]{FF9D9A}0.875 & \cellcolor[HTML]{FF9D9A}24.62 & \cellcolor[HTML]{FF9D9A}0.245 & 71           & \multicolumn{1}{c|}{112} &  %\\ \hline
%\multicolumn{1}{l|}{ } &   &            &  &   &  & 
%\multicolumn{1}{l|}{SemFeatSplat -- 32 (Ours)} & \textbf{0.875} & 24.64          & \textbf{0.244} & 56           & \multicolumn{1}{c|}{113} & 
\end{tabular}
\end{center}
\caption{Features Splatting evaluation on 21 scenes from ScanNet++}
\label{tab:scannetpp}
\end{table}

\paragraph{Out of distribution.} In order to analyze points of view far from the training cameras, we sample four points of view from simple linear trajectories on outdoor scenes, see Figure~\ref{fig:gs_dataset_2}. Observe how, in \textit{Train}, 3DGS generates floating 3D Gaussians to model far away clouds, which obfuscate the visibility for closer views.
Observe instead in FeatSplat--32 that, as the camera gets closer to the train, the Gaussians that before were white, first switch to blue and finally also to green, to better represent the background sky and trees.
A similar effect can be observed on the \textit{Bicycle} scene in the right top corner of the image. 
Finally, the \textit{Treehill} data reveals how FeatSplat learned to adapt the Gaussians' brightness for slightly different positions of the camera while 3DGS did not.    
\paragraph{Light encoding.} Furthermore, we analyze the pixel and position embeddings impact on the MLP by individually evaluating different input values for each of them, see Figure~\ref{fig:light_change}. Apparently, the MLP learns to use these embeddings to codify global lightning information. As a consequence, our model allows to modify the scene lightning at inference time without any addition training.
\begin{figure}[h!]
\begin{center}
      \includegraphics[width=\linewidth]{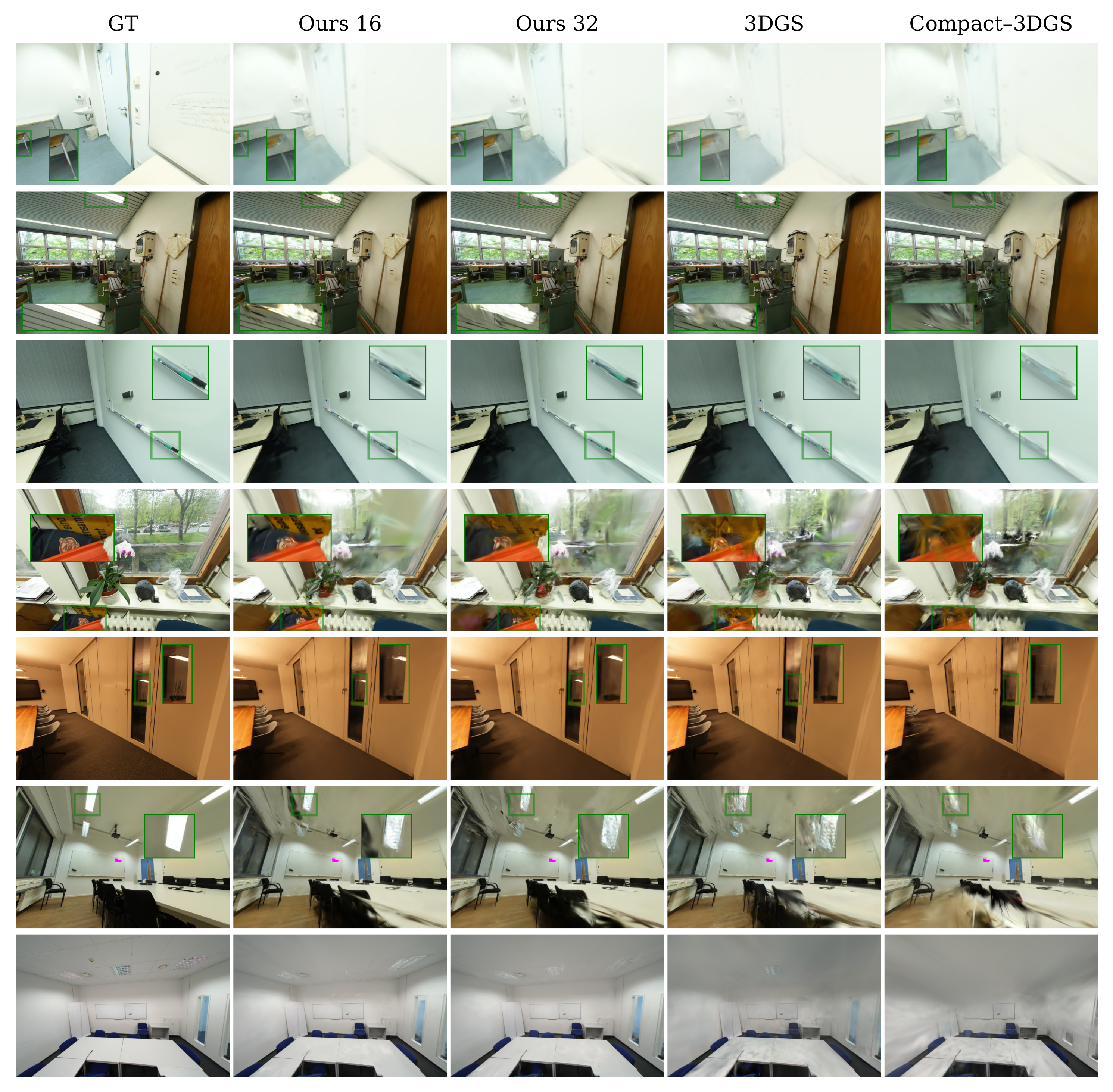}
\end{center}
\caption{Novel view synthesis of FeatSplat--16 and FeatSplat--32, against 3DGS~\cite{kerbl20233d} and Compact-3DGS~\cite{lee2023compact} on a subset of scenes from ScanNet++~\cite{yeshwanthliu2023scannetpp}.}
    \label{fig:scannetpp_dataset}
\end{figure}

\paragraph{ScanNet++.} Table~\ref{tab:scannetpp} shows that FeatSplat significantly outperforms both 3DGS and Compact-3DGS on all SSIM, PSNR, and LPIPS, with both FeatSplat--16 and FeatSplat--32 achieving the same performance.
The gap between FeatSplat and 3DGS is due to a more challenging test set, as previously mentioned. 
To properly learn view-dependent effects, 3DGS overfits to the training cameras. Some Gaussians are placed in such positions that are seen only in the optimized direction, while not being visible from other training directions without the same effect. This effect is unnoticed from points of view that are close to the training samples, or with a high visual overlap. However, for viewpoints far-away from the training ones, or with lower visual overlap, they become easily noticeable.
The ability of FeatSplat to encode multiple colors in the same Gaussian, and to condition the decoding on the point of view, reduces the impact of this problem. Figure~\ref{fig:scannetpp_dataset} shows how FeatSplat renders more accurate colors, captures smaller details, and reduces the number of artifacts.

\begin{figure}[h!]
\begin{center}
    \includegraphics[width=\linewidth]{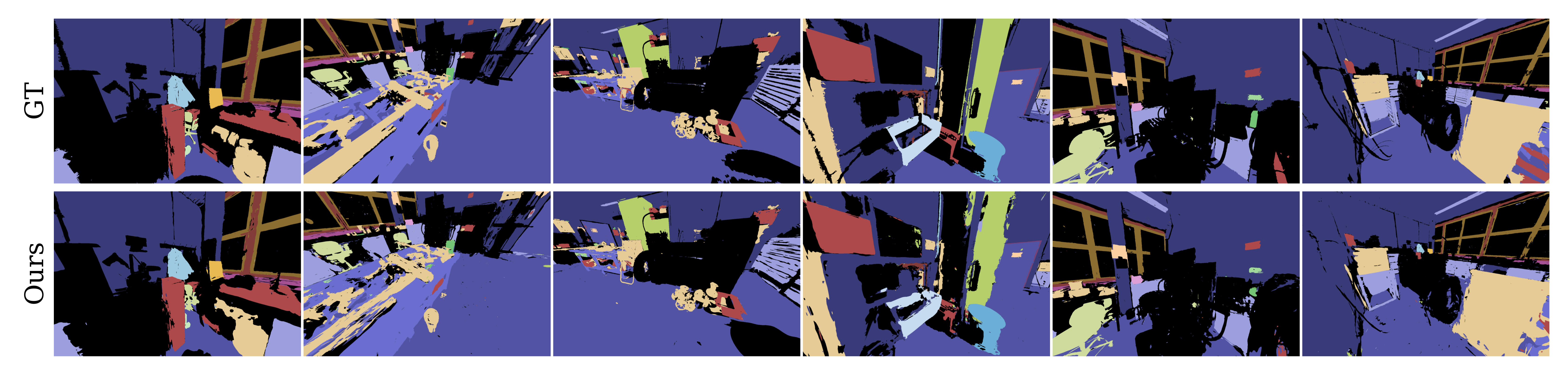}
\end{center}
\caption{Semantic FeatSplat with 32-dimensional vectors and 64 classes on ScanNet++.}
    \label{fig:scannetpp_dataset_2}
\end{figure}

\paragraph{Semantic FeatSplat} Our results so far suggest that 16 dimensions suffice to encode color. We then chose FeatSplat--32 to encode both RGB and semantic labels and perform semantic segmentation.
FeatSplat's segmentation masks for novel views achieve weighted mIOU of \(0.629\), maintaining the rendering performance with a PSNR of \(24.64\), and SSIM of \(0.875\) and an LPIPS of \(0.244\), and still rendering in real-time at 56 FPS.
The segmentation maps generated, see Figure~\ref{fig:scannetpp_dataset_2}, show that the model produces coherent segmentations even for small elements. However, the edges of the masks are noisy, as they follow Gaussians' shapes. Nevertheless, it proves the versatility of these features and their potential to encode different kind of information.
\subsection{Limitations} 
The small size of our MLP allows real-time rendering but limits the complexity of textures encoded in each Gaussian. Therefore, our FeatSplat still generates small artifacts in unobserved areas. Furthermore, our model sometimes over-smooths some highly textured areas, like the grass or some tree leaves. This trade-off between capacity and speed can have a negative impact in applications focused on rendering around the training cameras, for which generalization is less relevant. Finally, FeatSplat is slower than 3DGS in part due to the CUDA kernel implementation designed for RGB channels, with rendering time scaling linearly with the dimension of the alpha-blended feature vector. Nevertheless, the increased capacity of FeatSplat vectors halves the number of Gaussians per scene, limiting the training and rendering slow-down to just a half of 3DGS at a resolution of \(1752\times1168\).
\section{Conclusions}
In this paper we showed how the use of feature vectors to encode color information surpasses the performance of spherical harmonics when linked to small 3D primitives such as Gaussians for novel view synthesis. 
We argue that SHs lack capacity, and as a consequence an overpopulation of 3D Gaussians is required in order to approximate complex textures, which in turns results in the model overfitting the training viewpoints.
Our FeatSplat overcomes this issue by using feature vectors, alpha blended, concatenated to camera parameters and decoded by an MLP, that render a higher variety of more complex textures and hence reduce the Gaussians' overpopulation. 
FeatSplat outperforms previous methods on standard datasets in the literature and on the challenging ScanNet++~\cite{yeshwanthliu2023scannetpp}. In addition, we keep a competitive real-time rendering and enable the extension to other tasks. Finally, it is worth mentioning that FeatSplat should also benefit from any 3DGS optimizations, like the quantization proposed by Compact-3DGS~\cite{lee2023compact}, which would speed up its rendering.

\bibliography{main.bib}

\begin{thebibliography}{46}
\providecommand{\natexlab}[1]{#1}
\providecommand{\url}[1]{\texttt{#1}}
\expandafter\ifx\csname urlstyle\endcsname\relax
  \providecommand{\doi}[1]{doi: #1}\else
  \providecommand{\doi}{doi: \begingroup \urlstyle{rm}\Url}\fi

\bibitem[Barron et~al.(2022)Barron, Mildenhall, Verbin, Srinivasan, and
  Hedman]{barron2022mipnerf360}
Jonathan~T. Barron, Ben Mildenhall, Dor Verbin, Pratul~P. Srinivasan, and Peter
  Hedman.
\newblock Mip-nerf 360: Unbounded anti-aliased neural radiance fields.
\newblock \emph{CVPR}, 2022.

\bibitem[Deng et~al.(2022)Deng, Liu, Zhu, and Ramanan]{deng2022depth}
Kangle Deng, Andrew Liu, Jun-Yan Zhu, and Deva Ramanan.
\newblock Depth-supervised nerf: Fewer views and faster training for free.
\newblock In \emph{Proceedings of the IEEE/CVF Conference on Computer Vision
  and Pattern Recognition}, pages 12882--12891, 2022.

\bibitem[Fridovich-Keil et~al.(2022)Fridovich-Keil, Yu, Tancik, Chen, Recht,
  and Kanazawa]{fridovich2022plenoxels}
Sara Fridovich-Keil, Alex Yu, Matthew Tancik, Qinhong Chen, Benjamin Recht, and
  Angjoo Kanazawa.
\newblock Plenoxels: Radiance fields without neural networks.
\newblock In \emph{Proceedings of the IEEE/CVF Conference on Computer Vision
  and Pattern Recognition}, pages 5501--5510, 2022.

\bibitem[Guizilini et~al.(2022)Guizilini, Vasiljevic, Fang, Ambru,
  Shakhnarovich, Walter, and Gaidon]{guizilini2022depth}
Vitor Guizilini, Igor Vasiljevic, Jiading Fang, Rare Ambru, Greg Shakhnarovich,
  Matthew~R Walter, and Adrien Gaidon.
\newblock Depth field networks for generalizable multi-view scene
  representation.
\newblock In \emph{European Conference on Computer Vision}, pages 245--262.
  Springer, 2022.

\bibitem[Guizilini et~al.(2023{\natexlab{a}})Guizilini, Vasiljevic, Chen,
  Ambruș, and Gaidon]{guizilini2023towards}
Vitor Guizilini, Igor Vasiljevic, Dian Chen, Rareș Ambruș, and Adrien Gaidon.
\newblock Towards zero-shot scale-aware monocular depth estimation.
\newblock In \emph{Proceedings of the IEEE/CVF International Conference on
  Computer Vision}, pages 9233--9243, 2023{\natexlab{a}}.

\bibitem[Guizilini et~al.(2023{\natexlab{b}})Guizilini, Vasiljevic, Fang,
  Ambrus, Zakharov, Sitzmann, and Gaidon]{guizilini2023delira}
Vitor Guizilini, Igor Vasiljevic, Jiading Fang, Rares Ambrus, Sergey Zakharov,
  Vincent Sitzmann, and Adrien Gaidon.
\newblock Delira: Self-supervised depth, light, and radiance fields.
\newblock In \emph{Proceedings of the IEEE/CVF International Conference on
  Computer Vision}, pages 17935--17945, 2023{\natexlab{b}}.

\bibitem[Hedman et~al.(2018)Hedman, Philip, Price, Frahm, Drettakis, and
  Brostow]{hedman2018deep}
Peter Hedman, Julien Philip, True Price, Jan-Michael Frahm, George Drettakis,
  and Gabriel Brostow.
\newblock Deep blending for free-viewpoint image-based rendering.
\newblock \emph{ACM Transactions on Graphics (ToG)}, 37\penalty0 (6):\penalty0
  1--15, 2018.

\bibitem[Hendrycks and Gimpel(2023)]{hendrycks2023gaussian}
Dan Hendrycks and Kevin Gimpel.
\newblock Gaussian error linear units (gelus), 2023.

\bibitem[Kerbl et~al.(2023)Kerbl, Kopanas, Leimk{\"u}hler, and
  Drettakis]{kerbl20233d}
Bernhard Kerbl, Georgios Kopanas, Thomas Leimk{\"u}hler, and George Drettakis.
\newblock 3d gaussian splatting for real-time radiance field rendering.
\newblock \emph{ACM Transactions on Graphics}, 42\penalty0 (4):\penalty0 1--14,
  2023.

\bibitem[Kerr et~al.(2023)Kerr, Kim, Goldberg, Kanazawa, and
  Tancik]{kerr2023lerf}
Justin Kerr, Chung~Min Kim, Ken Goldberg, Angjoo Kanazawa, and Matthew Tancik.
\newblock Lerf: Language embedded radiance fields.
\newblock In \emph{Proceedings of the IEEE/CVF International Conference on
  Computer Vision}, pages 19729--19739, 2023.

\bibitem[Kingma and Ba(2015)]{Kingma2014AdamAM}
Diederik~P. Kingma and Jimmy Ba.
\newblock Adam: a method for stochastic optimization.
\newblock In \emph{International Conference for Learning Representations,},
  2015.

\bibitem[Kirillov et~al.(2023)Kirillov, Mintun, Ravi, Mao, Rolland, Gustafson,
  Xiao, Whitehead, Berg, Lo, Doll{\'a}r, and Girshick]{kirillov2023segany}
Alexander Kirillov, Eric Mintun, Nikhila Ravi, Hanzi Mao, Chloe Rolland, Laura
  Gustafson, Tete Xiao, Spencer Whitehead, Alexander~C. Berg, Wan-Yen Lo, Piotr
  Doll{\'a}r, and Ross Girshick.
\newblock Segment anything.
\newblock \emph{arXiv:2304.02643}, 2023.

\bibitem[Knapitsch et~al.(2017)Knapitsch, Park, Zhou, and
  Koltun]{knapitsch2017tanks}
Arno Knapitsch, Jaesik Park, Qian-Yi Zhou, and Vladlen Koltun.
\newblock Tanks and temples: Benchmarking large-scale scene reconstruction.
\newblock \emph{ACM Transactions on Graphics (ToG)}, 36\penalty0 (4):\penalty0
  1--13, 2017.

\bibitem[Kopanas et~al.(2021)Kopanas, Philip, Leimk{\"u}hler, and
  Drettakis]{kopanas2021point}
Georgios Kopanas, Julien Philip, Thomas Leimk{\"u}hler, and George Drettakis.
\newblock Point-based neural rendering with per-view optimization.
\newblock In \emph{Computer Graphics Forum}, volume~40, pages 29--43. Wiley
  Online Library, 2021.

\bibitem[Kopanas et~al.(2022)Kopanas, Leimk{\"u}hler, Rainer, Jambon, and
  Drettakis]{kopanas2022neural}
Georgios Kopanas, Thomas Leimk{\"u}hler, Gilles Rainer, Cl{\'e}ment Jambon, and
  George Drettakis.
\newblock Neural point catacaustics for novel-view synthesis of reflections.
\newblock \emph{ACM Transactions on Graphics (TOG)}, 41\penalty0 (6):\penalty0
  1--15, 2022.

\bibitem[Kundu et~al.(2022)Kundu, Genova, Yin, Fathi, Pantofaru, Guibas,
  Tagliasacchi, Dellaert, and Funkhouser]{kundu2022panoptic}
Abhijit Kundu, Kyle Genova, Xiaoqi Yin, Alireza Fathi, Caroline Pantofaru,
  Leonidas~J Guibas, Andrea Tagliasacchi, Frank Dellaert, and Thomas
  Funkhouser.
\newblock Panoptic neural fields: A semantic object-aware neural scene
  representation.
\newblock In \emph{Proceedings of the IEEE/CVF Conference on Computer Vision
  and Pattern Recognition}, pages 12871--12881, 2022.

\bibitem[Lee et~al.(2023)Lee, Rho, Sun, Ko, and Park]{lee2023compact}
Joo~Chan Lee, Daniel Rho, Xiangyu Sun, Jong~Hwan Ko, and Eunbyung Park.
\newblock Compact 3d gaussian representation for radiance field.
\newblock \emph{arXiv preprint arXiv:2311.13681}, 2023.

\bibitem[Li et~al.(2021)Li, Feng, She, Ding, Wang, and Lee]{li2021mine}
Jiaxin Li, Zijian Feng, Qi~She, Henghui Ding, Changhu Wang, and Gim~Hee Lee.
\newblock Mine: Towards continuous depth mpi with nerf for novel view
  synthesis.
\newblock In \emph{Proceedings of the IEEE/CVF International Conference on
  Computer Vision}, pages 12578--12588, 2021.

\bibitem[Liu et~al.(2020)Liu, Gu, Zaw~Lin, Chua, and Theobalt]{liu2020neural}
Lingjie Liu, Jiatao Gu, Kyaw Zaw~Lin, Tat-Seng Chua, and Christian Theobalt.
\newblock Neural sparse voxel fields.
\newblock \emph{Advances in Neural Information Processing Systems},
  33:\penalty0 15651--15663, 2020.

\bibitem[Merrill and Grimshaw(2010)]{merrill2010revisiting}
Duane~G Merrill and Andrew~S Grimshaw.
\newblock Revisiting sorting for gpgpu stream architectures.
\newblock In \emph{Proceedings of the 19th international conference on Parallel
  architectures and compilation techniques}, pages 545--546, 2010.

\bibitem[Mildenhall et~al.(2021)Mildenhall, Srinivasan, Tancik, Barron,
  Ramamoorthi, and Ng]{mildenhall2021nerf}
Ben Mildenhall, Pratul~P Srinivasan, Matthew Tancik, Jonathan~T Barron, Ravi
  Ramamoorthi, and Ren Ng.
\newblock Nerf: Representing scenes as neural radiance fields for view
  synthesis.
\newblock \emph{Communications of the ACM}, 65\penalty0 (1):\penalty0 99--106,
  2021.

\bibitem[M{\"u}ller et~al.(2022)M{\"u}ller, Evans, Schied, and
  Keller]{muller2022instant}
Thomas M{\"u}ller, Alex Evans, Christoph Schied, and Alexander Keller.
\newblock Instant neural graphics primitives with a multiresolution hash
  encoding.
\newblock \emph{ACM transactions on graphics (TOG)}, 41\penalty0 (4):\penalty0
  1--15, 2022.

\bibitem[Neff et~al.(2021)Neff, Stadlbauer, Parger, Kurz, Mueller, Chaitanya,
  Kaplanyan, and Steinberger]{neff2021donerf}
Thomas Neff, Pascal Stadlbauer, Mathias Parger, Andreas Kurz, Joerg~H Mueller,
  Chakravarty R~Alla Chaitanya, Anton Kaplanyan, and Markus Steinberger.
\newblock Donerf: Towards real-time rendering of compact neural radiance fields
  using depth oracle networks.
\newblock In \emph{Computer Graphics Forum}, volume~40, pages 45--59. Wiley
  Online Library, 2021.

\bibitem[Oquab et~al.(2023)Oquab, Darcet, Moutakanni, Vo, Szafraniec, Khalidov,
  Fernandez, Haziza, Massa, El-Nouby, Howes, Huang, Xu, Sharma, Li, Galuba,
  Rabbat, Assran, Ballas, Synnaeve, Misra, Jegou, Mairal, Labatut, Joulin, and
  Bojanowski]{oquab2023dinov2}
Maxime Oquab, Timothée Darcet, Theo Moutakanni, Huy~V. Vo, Marc Szafraniec,
  Vasil Khalidov, Pierre Fernandez, Daniel Haziza, Francisco Massa, Alaaeldin
  El-Nouby, Russell Howes, Po-Yao Huang, Hu~Xu, Vasu Sharma, Shang-Wen Li,
  Wojciech Galuba, Mike Rabbat, Mido Assran, Nicolas Ballas, Gabriel Synnaeve,
  Ishan Misra, Herve Jegou, Julien Mairal, Patrick Labatut, Armand Joulin, and
  Piotr Bojanowski.
\newblock Dinov2: Learning robust visual features without supervision, 2023.

\bibitem[Qin et~al.(2023)Qin, Li, Zhou, Wang, and Pfister]{qin2023langsplat}
Minghan Qin, Wanhua Li, Jiawei Zhou, Haoqian Wang, and Hanspeter Pfister.
\newblock Langsplat: 3d language gaussian splatting.
\newblock \emph{arXiv preprint arXiv:2312.16084}, 2023.

\bibitem[Radford et~al.(2021)Radford, Kim, Hallacy, Ramesh, Goh, Agarwal,
  Sastry, Askell, Mishkin, Clark, et~al.]{radford2021learning}
Alec Radford, Jong~Wook Kim, Chris Hallacy, Aditya Ramesh, Gabriel Goh,
  Sandhini Agarwal, Girish Sastry, Amanda Askell, Pamela Mishkin, Jack Clark,
  et~al.
\newblock Learning transferable visual models from natural language
  supervision.
\newblock In \emph{International conference on machine learning}, pages
  8748--8763. PMLR, 2021.

\bibitem[Roessle et~al.(2022)Roessle, Barron, Mildenhall, Srinivasan, and
  Nie{\ss}ner]{roessle2022dense}
Barbara Roessle, Jonathan~T Barron, Ben Mildenhall, Pratul~P Srinivasan, and
  Matthias Nie{\ss}ner.
\newblock Dense depth priors for neural radiance fields from sparse input
  views.
\newblock In \emph{Proceedings of the IEEE/CVF Conference on Computer Vision
  and Pattern Recognition}, pages 12892--12901, 2022.

\bibitem[Sandstr{\"o}m et~al.(2023)Sandstr{\"o}m, Li, Van~Gool, and
  Oswald]{sandstrom2023point}
Erik Sandstr{\"o}m, Yue Li, Luc Van~Gool, and Martin~R Oswald.
\newblock Point-slam: Dense neural point cloud-based slam.
\newblock In \emph{Proceedings of the IEEE/CVF International Conference on
  Computer Vision}, pages 18433--18444, 2023.

\bibitem[Shen et~al.(2023)Shen, Yang, Yu, Wong, Kaelbling, and
  Isola]{shen2023distilled}
William Shen, Ge~Yang, Alan Yu, Jansen Wong, Leslie~Pack Kaelbling, and Phillip
  Isola.
\newblock Distilled feature fields enable few-shot language-guided
  manipulation.
\newblock In \emph{Conference on Robot Learning}, pages 405--424. PMLR, 2023.

\bibitem[Sucar et~al.(2021)Sucar, Liu, Ortiz, and Davison]{sucar2021imap}
Edgar Sucar, Shikun Liu, Joseph Ortiz, and Andrew~J Davison.
\newblock imap: Implicit mapping and positioning in real-time.
\newblock In \emph{Proceedings of the IEEE/CVF International Conference on
  Computer Vision}, pages 6229--6238, 2021.

\bibitem[WANG et~al.(2023)WANG, Chen, and Yang]{wang2023dmnerf}
Bing WANG, Lu~Chen, and Bo~Yang.
\newblock {DM}-ne{RF}: 3d scene geometry decomposition and manipulation from 2d
  images.
\newblock In \emph{The Eleventh International Conference on Learning
  Representations}, 2023.

\bibitem[Wang et~al.(2021)Wang, Liu, Liu, Theobalt, Komura, and
  Wang]{wang2021neus}
Peng Wang, Lingjie Liu, Yuan Liu, Christian Theobalt, Taku Komura, and Wenping
  Wang.
\newblock Neus: Learning neural implicit surfaces by volume rendering for
  multi-view reconstruction.
\newblock \emph{Advances in Neural Information Processing Systems},
  34:\penalty0 27171--27183, 2021.

\bibitem[Wang et~al.(2023)Wang, Han, Habermann, Daniilidis, Theobalt, and
  Liu]{neus2}
Yiming Wang, Qin Han, Marc Habermann, Kostas Daniilidis, Christian Theobalt,
  and Lingjie Liu.
\newblock Neus2: Fast learning of neural implicit surfaces for multi-view
  reconstruction.
\newblock In \emph{Proceedings of the IEEE/CVF International Conference on
  Computer Vision (ICCV)}, 2023.

\bibitem[Wang et~al.(2004)Wang, Bovik, Sheikh, and Simoncelli]{ssim}
Zhou Wang, A.C. Bovik, H.R. Sheikh, and E.P. Simoncelli.
\newblock Image quality assessment: from error visibility to structural
  similarity.
\newblock \emph{IEEE Transactions on Image Processing}, 13\penalty0
  (4):\penalty0 600--612, 2004.
\newblock \doi{10.1109/TIP.2003.819861}.

\bibitem[Wei et~al.(2021)Wei, Liu, Rao, Zhao, Lu, and Zhou]{wei2021nerfingmvs}
Yi~Wei, Shaohui Liu, Yongming Rao, Wang Zhao, Jiwen Lu, and Jie Zhou.
\newblock Nerfingmvs: Guided optimization of neural radiance fields for indoor
  multi-view stereo.
\newblock In \emph{Proceedings of the IEEE/CVF International Conference on
  Computer Vision}, pages 5610--5619, 2021.

\bibitem[Xu et~al.(2022)Xu, Xu, Philip, Bi, Shu, Sunkavalli, and
  Neumann]{xu2022point}
Qiangeng Xu, Zexiang Xu, Julien Philip, Sai Bi, Zhixin Shu, Kalyan Sunkavalli,
  and Ulrich Neumann.
\newblock Point-nerf: Point-based neural radiance fields.
\newblock In \emph{Proceedings of the IEEE/CVF conference on computer vision
  and pattern recognition}, pages 5438--5448, 2022.

\bibitem[Ye et~al.(2024)Ye, Danelljan, Yu, and Ke]{gaussian_grouping}
Mingqiao Ye, Martin Danelljan, Fisher Yu, and Lei Ke.
\newblock Gaussian grouping: Segment and edit anything in 3d scenes.
\newblock In \emph{European Conference on Computer Vision}. Springer, 2024.

\bibitem[Yeshwanth et~al.(2023)Yeshwanth, Liu, Nie{\ss}ner, and
  Dai]{yeshwanthliu2023scannetpp}
Chandan Yeshwanth, Yueh-Cheng Liu, Matthias Nie{\ss}ner, and Angela Dai.
\newblock Scannet++: A high-fidelity dataset of 3d indoor scenes.
\newblock In \emph{Proceedings of the International Conference on Computer
  Vision ({ICCV})}, 2023.

\bibitem[Yifan et~al.(2019)Yifan, Serena, Wu, {\"O}ztireli, and
  Sorkine-Hornung]{yifan2019differentiable}
Wang Yifan, Felice Serena, Shihao Wu, Cengiz {\"O}ztireli, and Olga
  Sorkine-Hornung.
\newblock Differentiable surface splatting for point-based geometry processing.
\newblock \emph{ACM Transactions on Graphics (TOG)}, 38\penalty0 (6):\penalty0
  1--14, 2019.

\bibitem[Yu et~al.(2021)Yu, Li, Tancik, Li, Ng, and
  Kanazawa]{yu2021plenoctrees}
Alex Yu, Ruilong Li, Matthew Tancik, Hao Li, Ren Ng, and Angjoo Kanazawa.
\newblock Plenoctrees for real-time rendering of neural radiance fields.
\newblock In \emph{Proceedings of the IEEE/CVF International Conference on
  Computer Vision}, pages 5752--5761, 2021.

\bibitem[Zhang et~al.(2021)Zhang, Liu, Ye, Zhao, Zhang, Wu, Zhang, Xu, and
  Yu]{zhang2021editable}
Jiakai Zhang, Xinhang Liu, Xinyi Ye, Fuqiang Zhao, Yanshun Zhang, Minye Wu,
  Yingliang Zhang, Lan Xu, and Jingyi Yu.
\newblock Editable free-viewpoint video using a layered neural representation.
\newblock \emph{ACM Transactions on Graphics}, 40\penalty0 (4):\penalty0 1--18,
  2021.

\bibitem[Zhang et~al.(2018)Zhang, Isola, Efros, Shechtman, and
  Wang]{Zhang_2018_CVPR}
Richard Zhang, Phillip Isola, Alexei~A. Efros, Eli Shechtman, and Oliver Wang.
\newblock The unreasonable effectiveness of deep features as a perceptual
  metric.
\newblock In \emph{Proceedings of the IEEE Conference on Computer Vision and
  Pattern Recognition (CVPR)}, June 2018.

\bibitem[Zhou et~al.(2024)Zhou, Chang, Jiang, Fan, Zhu, Xu, Chari, You, Wang,
  and Kadambi]{zhou2024feature}
Shijie Zhou, Haoran Chang, Sicheng Jiang, Zhiwen Fan, Zehao Zhu, Dejia Xu,
  Pradyumna Chari, Suya You, Zhangyang Wang, and Achuta Kadambi.
\newblock Feature 3dgs: Supercharging 3d gaussian splatting to enable distilled
  feature fields.
\newblock In \emph{Proceedings of the IEEE/CVF Conference on Computer Vision
  and Pattern Recognition}, pages 21676--21685, 2024.

\bibitem[Zhu et~al.(2022)Zhu, Peng, Larsson, Xu, Bao, Cui, Oswald, and
  Pollefeys]{zhu2022nice}
Zihan Zhu, Songyou Peng, Viktor Larsson, Weiwei Xu, Hujun Bao, Zhaopeng Cui,
  Martin~R Oswald, and Marc Pollefeys.
\newblock Nice-slam: Neural implicit scalable encoding for slam.
\newblock In \emph{Proceedings of the IEEE/CVF Conference on Computer Vision
  and Pattern Recognition}, pages 12786--12796, 2022.

\bibitem[Zuo et~al.(2024)Zuo, Samangouei, Zhou, Di, and Li]{zuo2024fmgs}
Xingxing Zuo, Pouya Samangouei, Yunwen Zhou, Yan Di, and Mingyang Li.
\newblock Fmgs: Foundation model embedded 3d gaussian splatting for holistic 3d
  scene understanding.
\newblock \emph{arXiv preprint arXiv:2401.01970}, 2024.

\bibitem[Zwicker et~al.(2001)Zwicker, Pfister, Van~Baar, and
  Gross]{zwicker2001ewa}
Matthias Zwicker, Hanspeter Pfister, Jeroen Van~Baar, and Markus Gross.
\newblock Ewa volume splatting.
\newblock In \emph{Proceedings Visualization, 2001. VIS'01.}, pages 29--538.
  IEEE, 2001.

\end{thebibliography}
\clearpage
\appendix

\section{Model details}
\label{app:implementation}
\subsection{Implementation}
We use a two-layer MLP with bias. The first layer has 64 neurons and a SiLU activation function~\cite{hendrycks2023gaussian}. For the RGB-only implementation, the output layers have 3 output neurons and a Sigmoid activation function. Instead, for the semantic segmentation implementation, the output layer has 67 output neurons, with a Sigmoid activation function for the first 3 channels, and a Softmax for the last 64. 
We use the Adam optimizer with \(\epsilon=1\times10^{-15}\), a learning rate of \(0.001\) for the MLP and \(0.0025\) for the feature vectors, and \(\lambda_{sem} = 0.001\) for the semantic segmentation model. The rest of the Gaussians' parameters and optimization hyperparameters follow 3DGS's~\cite{kerbl20233d} original implementation.
We use the 3DGS~\cite{kerbl20233d} CUDA differentiable rasterizer. We adapt it to our model by skipping spherical harmonics and passing feature vectors as precomputed colors. To do it, we change the number of channels when compiling from 3 to 16 and 32 for FeatSplat-16 and FeatSplat-32 respectively.
The model has been implemented on Pytorch 1.12 and CUDA SDK 11.6.2.

\subsection{Embeddings ablation}
\label{sec:ablation}
We ablate the impact on the 32-dimensional model of concatenating to the pre-MLP feature vectors the 3D camera postion coordinates \(\mathbf{x}_{cam}\), its rotation using Euler angles \(\mathbf{x}_{rot}\), and the 2 dimension embedding of the relative position of each pixel \(\mathbf{e}_p\). The evaluation is done on the scenes \textit{Bicycle} from Mip-360~\cite{barron2022mipnerf360} and \textit{Train} from T\&T ~\cite{knapitsch2017tanks}.

\begin{table}[h]
\small
\begin{center}
\begin{tabular}{lcccccc}
\multicolumn{1}{c}{\textbf{}}                                      & \multicolumn{3}{c}{Bicycle}                                                               & \multicolumn{3}{c}{Train}                                                                   \\
\multicolumn{1}{l|}{FeatSplat -- 32}                                              & \multicolumn{1}{c}{SSIM} & \multicolumn{1}{c}{PSNR} & \multicolumn{1}{c|}{LPIPS}          & \multicolumn{1}{c}{SSIM} & \multicolumn{1}{c}{PSNR} & LPIPS            \\ \hline
\multicolumn{1}{l|}{w/o embeddings} & 0.7415                   & 24.54                    & \multicolumn{1}{l|}{0.241}          & 0.807                    & 22.02                    & 0.221            \\ \hline
\multicolumn{1}{l|}{only \(\mathbf{e}_p\)}                      & 0.745                    & 24.70                    & \multicolumn{1}{l|}{0.239}          & \textbf{0.815}           & 22.84                    & 0.216            \\
\multicolumn{1}{l|}{only \(\mathbf{x}_{cam}\)}                          & 0.748                    & 24.84                    & \multicolumn{1}{l|}{0.239}          & 0.808                    & 22.09                    & 0.219            \\
\multicolumn{1}{l|}{only \(\mathbf{x}_{rot}\)}                      & 0.743                    & 24.52                    & \multicolumn{1}{l|}{0.240}          & 0.810           & 22.03                    & 0.22            \\ \hline
\multicolumn{1}{l|}{with  \(\mathbf{e}_p\) and \(\mathbf{x}_{rot}\)}                          & 0.746                    & 24.69                    & \multicolumn{1}{l|}{0.239}          & 0.813                    & 22.748                    & 0.215            \\
\multicolumn{1}{l|}{with \(\mathbf{x}_{cam}\) and \(\mathbf{x}_{rot}\)}                      & 0.748                    & 24.75                    & \multicolumn{1}{l|}{0.240}          & 0.813           & 22.58                    & 0.216     \\      
\multicolumn{1}{l|}{\textbf{with} \(\mathbf{e}_p\) \textbf{and} \(\mathbf{x}_{cam}\)}                      & \textbf{0.751}           & \textbf{25.02}           & \multicolumn{1}{l|}{\textbf{0.237}} & \textbf{0.815}           & \textbf{22.85}           & \textbf{0.215}            \\ \hline

\multicolumn{1}{l|}{with \(\mathbf{e}_p\), \(\mathbf{x}_{cam}\) and \(\mathbf{x}_{rot}\)}                             & 0.749           & 25.01           & \multicolumn{1}{l|}{0.238} & 0.813           & 22.84           & 0.216  \\
\end{tabular}
\end{center}
\caption{FeatSplat embbedings ablation on scenes \textit{Bicycle} from Mip-360~\cite{barron2022mipnerf360} and \textit{Train} from T\&T ~\cite{knapitsch2017tanks}. We integrate in the final model only \(\mathbf{e}_p\) and \(\mathbf{x}_{cam}\).}
\label{tab:ablation}
\end{table}
The ablation results in Table~\ref{tab:ablation} show that both \(\mathbf{e}_p\) and \(\mathbf{x}_{cam}\) have by themselves a positive impact on the metrics, while \(\mathbf{x}_{cam}\) does not improve against not adding it. 
While \(\mathbf{e}_p\) is responsible from almost all the improvement on \textit{Train}, \(\mathbf{x}_{cam}\) has a better performance on \textit{Bicycle}, and both together achieve the best performance. 
On the other hand, \(\mathbf{x}_{rot}\) alone does not improve the performance. Combining it with \(\mathbf{x}_{cam}\) improves the performance of \(\mathbf{x}_{cam}\) alone on \textit{Train}, although less than \(\mathbf{e}_p\). 
The results show that \(\mathbf{x}_{rot}\) is redundant, with \(\mathbf{e}_p\) and \(\mathbf{x}_{cam}\) having the best performance.

\section{ScanNet++ Novel View Synthesis Benchmark}

\begin{table}[h!]
\small
\begin{center}
\begin{tabular}{lccccc}\multicolumn{1}{c}{\textbf{}}                   & \multicolumn{4}{c}{ScanNet++}                                                                       &  \\
\multicolumn{1}{l|}{}                           & SSIM           & PSNR           & LPIPS          & \multicolumn{1}{c|}{FPS}          &  \\ \hline
\multicolumn{1}{l|}{HW GTS - RPBG}      & \cellcolor[HTML]{FF9D9A}0.873          & \cellcolor[HTML]{FF9D9A} 24.355          & \cellcolor[HTML]{FF9D9A} 0.280   &  \multicolumn{1}{c|}{\(\thicksim 1\)} & \\
\multicolumn{1}{l|}{\textbf{FeatSplat--32 (Ours)}}          & \cellcolor[HTML]{FFFC9E}0.869 & \cellcolor[HTML]{FFCE93}24.247 & \cellcolor[HTML]{FFCE93}0.314 &   \multicolumn{1}{c|}{\cellcolor[HTML]{FFCE93}50} & \\ %\\ \hline
\multicolumn{1}{l|}{Nerfacto}          & 0.861 & 24.049 & 0.342 & \multicolumn{1}{c|}{0.3} & \\ %\\ \hline
\multicolumn{1}{l|}{TensoRF}          & 0.849 & 23.978 & 0.407 & \multicolumn{1}{c|}{--} & \\ %\\ \hline
\multicolumn{1}{l|}{3D Gaussian Splatting}          & \cellcolor[HTML]{FFCE93}0.871 & \cellcolor[HTML]{FFFC9E}23.891 & \cellcolor[HTML]{FFFC9E}0.319 &  \multicolumn{1}{c|}{\cellcolor[HTML]{FF9D9A}155} & \\ %\\ \hline
\multicolumn{1}{l|}{iNGP}          & 0.859 & 23.812 & 0.375 &  \multicolumn{1}{c|}{0.3} & \\ %\\ \hline
\end{tabular}
\end{center}
\caption{ScanNet++ Novel View Synthesis Benchmark}
\label{tab:scannetpp_bench}
\end{table}

\section{Datasets}
\label{app:datasets}
\subsection{Mip\-360, T\&T and DB.} The three typical datasets we use amount to a total of 12 scenes: Mip\-360\cite{barron2022mipnerf360}, with 9 scenes, 5 outdoor at \(1237\times822\) (\textit{Bicycle}, \textit{Flowers}, \textit{Garden}, \textit{Stump}, \textit{Treehill},) and 4 indoor at \(1558\times1039\) (\textit{Bonsai}, \textit{Counter}, \textit{Kitchen}, \textit{Room}), with an average of 216 views per scene; Tanks and Temples~\cite{knapitsch2017tanks} with 2 outdoor scenes at \(979\times546\) (Train and Truck) and an average of 276 views per scene; and Deep Blending~\cite{hedman2018deep} with 2 indoor scenes (\textit{DrJohnson} and \textit{Playroom}) at \(1264\times832\) and an average of 244 views per scene. In these datasets the testing set is selected sampling one every eight frames of the camera trajectory, using the other 7 frames to train.

\subsection{ScanNet++}
ScanNet++ is large-scale dataset with 460 indoor scenes and high-quality RGB and semantic annotations. 
It has been designed for evaluating high resolution novel view synthesis, with a resolution of undistorted images of \(1752\times1168\) and a test set generated from an independent camera trajectory.
The subset of used scenes has been randomly selected ensuring the fit in memory 24 gigabytes of vRAM in a single RTX 3090 GPU as Float-Point 32. As a consequence, the selected subset of scenes has an average of 408 views per scene, higher than the others datasets used.
The evaluated scenes are: 
\textit{0a5c013435}, \textit{f07340dfea}, \textit{7bc286c1b6}, \textit{d2f44bf242}, \textit{85251de7d1}, \textit{0e75f3c4d9}, \textit{98fe276aa8}, \textit{7e7cd69a59}, \textit{f3685d06a9}, \textit{21d970d8de}, \textit{8b5caf3398}, \textit{ada5304e41}, \textit{4c5c60fa76}, \textit{ebc200e928}, \textit{a5114ca13d}, \textit{5942004064}, \textit{1ada7a0617}, \textit{f6659a3107}, \textit{1a130d092a}, \textit{80ffca8a48}, \textit{08bbbdcc3d},
\label{app:scannetpp}
\subsection{Test and Train cameras plots}
\begin{figure}[h]
    \centering
    \includegraphics[width=0.89\linewidth]{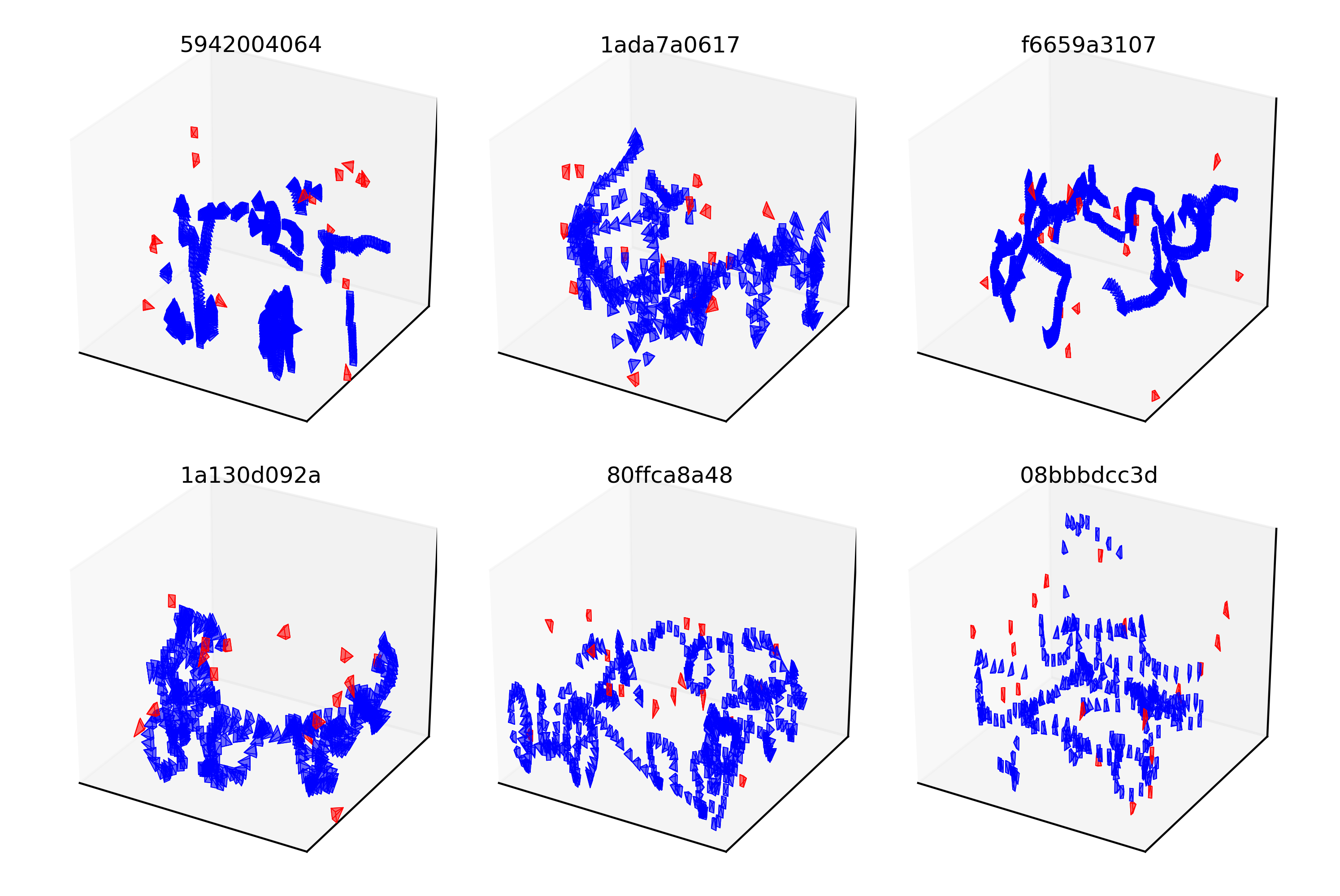}
    \caption{Train (blue) and test (red) and camera for ScanNet++}
    \label{fig:app_scannet_pp_1}
\end{figure}

\begin{figure}
    \centering
    \includegraphics[width=\linewidth]{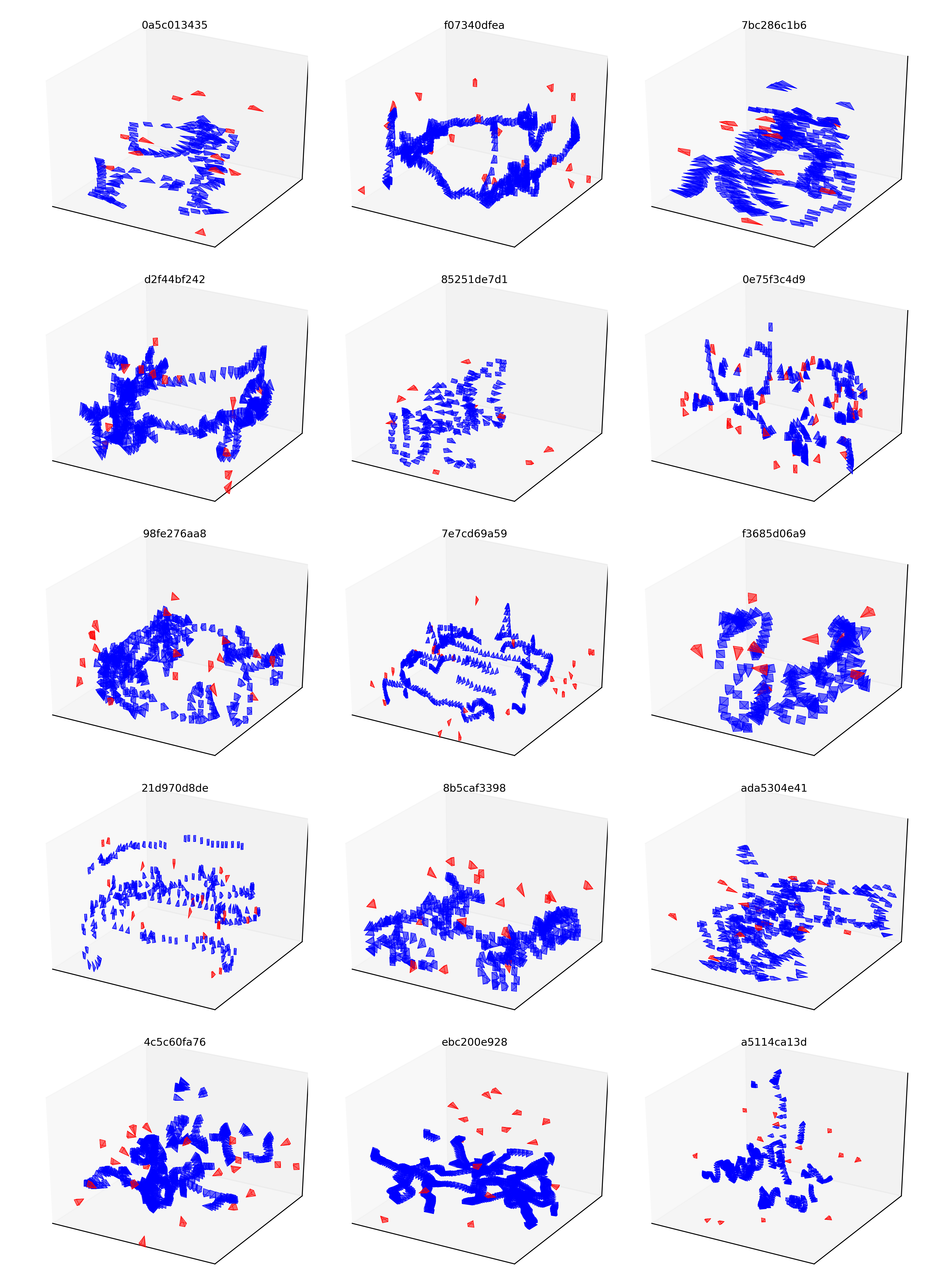}
    \caption{Train (blue) and test (red) and camera for ScanNet++}
    \label{fig:app_scannet_pp_0}
\end{figure}

\begin{figure}
    \centering
    \includegraphics[width=\linewidth]{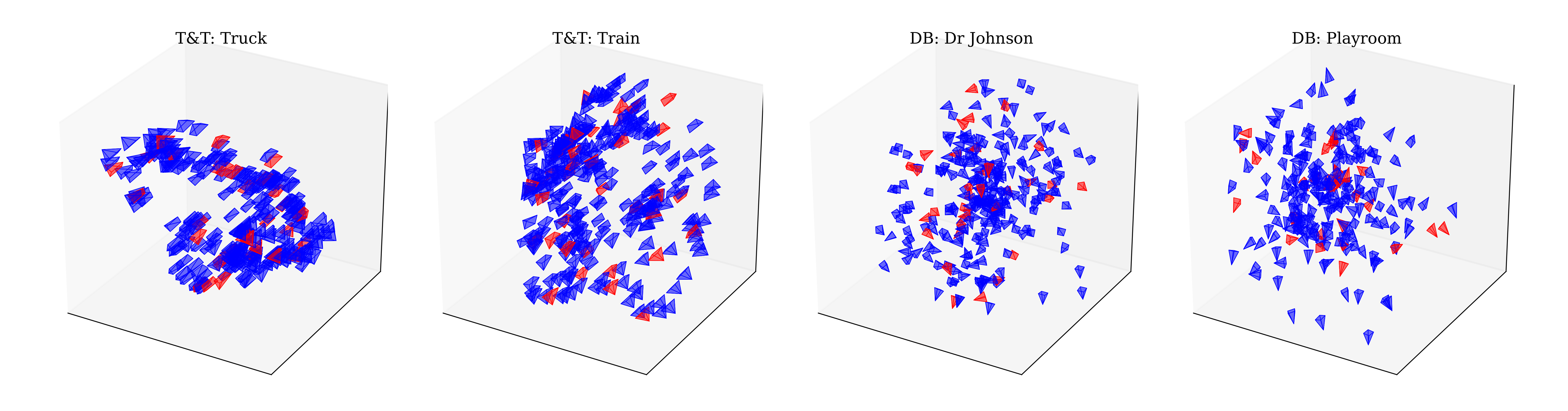}
    \caption{Train (blue) and test (red) and camera for T\&T and DB}
    \label{fig:app_tandt_db}
\end{figure}

\begin{figure}
    \centering
    \includegraphics[width=\linewidth]{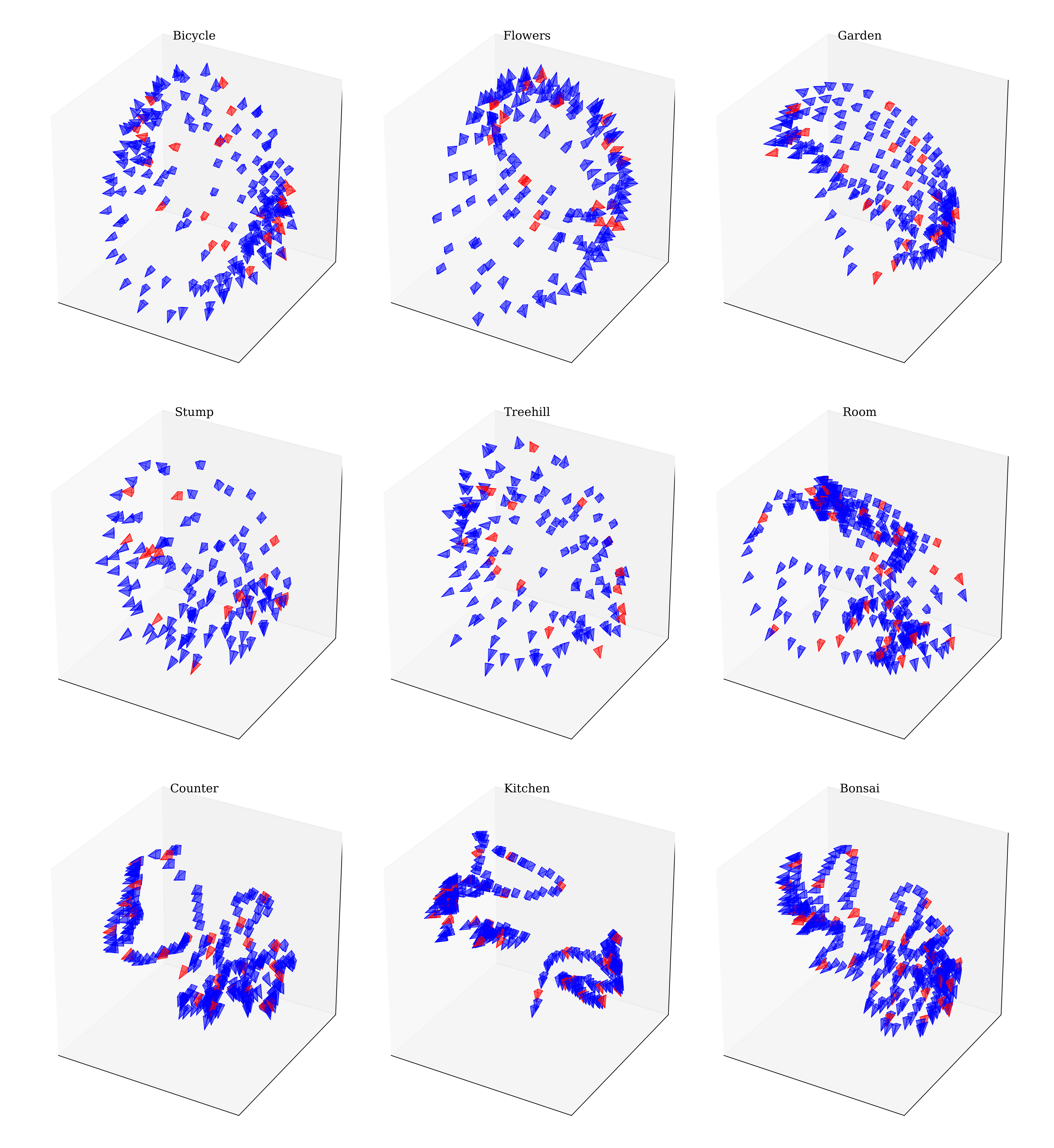}
    \caption{Train (blue) and test (red) and camera for Mip-360}
    \label{fig:app_mip360}
\end{figure}

Figures~\ref{fig:app_scannet_pp_0},~\ref{fig:app_scannet_pp_1},~\ref{fig:app_tandt_db} and~\ref{fig:app_mip360} show the distribution of train and test cameras for all evaluated scenes.


\end{document}